\documentclass[10pt, conference, compsoc]{IEEEtran}

\usepackage{cite}
\usepackage{amsmath,amssymb,amsfonts}
\usepackage{algorithmic}
\usepackage{graphicx}
\usepackage{textcomp}
\usepackage{booktabs}
\usepackage{colortbl}
\usepackage{multirow}
\usepackage{hhline}
\usepackage{float}

\usepackage{color}
\usepackage{subcaption}
\usepackage{hyperref}
\usepackage{siunitx}

\usepackage{stfloats}

\usepackage[capitalize]{cleveref}
\crefname{section}{Sec.}{Secs.}
\Crefname{section}{Section}{Sections}
\Crefname{table}{Table}{Tables}
\crefname{table}{Tab.}{Tabs.}

\newcommand{\ie}{i.e., }
\newcommand{\eg}{e.g., }
\newcommand{\etal}{\emph{et al.\ }}

\newcommand{\diff}[1]{\textcolor{red}{($#1$)}}
\makeatletter
\newcommand\notsotiny{\@setfontsize\notsotiny{6}{7}}
\makeatother
\definecolor{grady-dark}{gray}{0.9}
\definecolor{grady-med}{gray}{0.95}

\begin{document}
\title{Multi-scale Feature Alignment for\\Continual Learning of Unlabeled Domains}

\author{
\IEEEauthorblockN{Kevin Thandiackal$^{1,2*}$,
Luigi Piccinelli$^{3*\dagger}$,
Pushpak Pati$^{1}$ and
Orcun Goksel$^{2,4}$}\\
\IEEEauthorblockA{$^{1}$IBM Research Europe, Zurich, Switzerland}
\IEEEauthorblockA{$^{2}$Computer-assisted Applications in Medicine, ETH Zurich, Zurich, Switzerland}
\IEEEauthorblockA{$^{3}$Visual Intelligence and Systems, ETH Zurich, Zurich, Switzerland}
\IEEEauthorblockA{$^{4}$Department of Information Technology, Uppsala University, Uppsala, Sweden}
}

\maketitle

\makeatletter%
\long\def\@makefntext#1{%
  \noindent \hb@xt@ 1.8em{\hss\@makefnmark}#1}
\makeatother

\begingroup\renewcommand\thefootnote{*}
\footnotetext{The authors contributed equally to this work.}
\endgroup
\begingroup\renewcommand\thefootnote{$\dagger$}
\footnotetext{Work done while at IBM Research Europe.}
\endgroup

\begin{abstract}
Methods for unsupervised domain adaptation (UDA) help to improve the performance of deep neural networks on unseen domains without any labeled data.
Especially in medical disciplines such as histopathology, this is crucial since large datasets with detailed annotations are scarce.
While the majority of existing UDA methods focus on the adaptation from a labeled source to a single unlabeled target domain, many real-world applications with a long life cycle involve more than one target domain.
Thus, the ability to sequentially adapt to multiple target domains becomes essential.
In settings where the data from previously seen domains cannot be stored, \eg due to data protection regulations, the above becomes a challenging continual learning problem.
To this end, we propose to use generative feature-driven image replay in conjunction with a dual-purpose discriminator that not only enables the generation of images with realistic features for replay, but also promotes feature alignment during domain adaptation.
We evaluate our approach extensively on a sequence of three histopathological datasets for tissue-type classification, achieving state-of-the-art results.
We present detailed ablation experiments studying our proposed method components and demonstrate a possible use-case of our continual UDA method for an unsupervised patch-based segmentation task given high-resolution tissue images.
\end{abstract}


\IEEEpeerreviewmaketitle

\section{Introduction}
\label{sec:introduction}

Histopathological analysis is the gold standard for cancer diagnosis.
Recent technological advancements have enabled digitization of entire histological slides, thus allowing pathologists to perform remote diagnoses on computers. 
The adoption of digital workflows has also empowered the collection of large-scale data and facilitated the development of deep learning methods for various computational pathology (CP) tasks~\cite{campanella2019clinical,lu2021clam,shao2021transmil,graham2019hovernet,pati2022hactnet,thandiackal2022differentiable,pati2023weakly}.
However, several shortcomings still limit their applicability in practice.
First, many CP methods perform poorly on data that comes from a different domain than the training domain~\cite{vanderlaak2021dlhisto}.
This is caused by domain shifts resulting from, \eg differences in data acquisition (scanners or staining), annotation protocols, or patient demographics.
Second, since most successful methods are based on deep learning, their training requires large annotated datasets.
Especially in healthcare, such annotations are exceedingly expensive as they require rigorous efforts and time from trained experts.
Hence, the development of unsupervised methods is essential to facilitate wide-spread adoption in practice.

Above-mentioned challenges are tackled by Unsupervised Domain Adaptation (UDA)~\cite{kouw2019udareview,wilson2020udasurvey}, which is concerned with adapting models trained on a labeled source domain to a target domain for which only unlabeled samples are available.
Most existing UDA methods focus on a setting with one source and one target dataset~\cite{ganin2016dann,tzeng2017adda,long2018cdan,xu2019afn,jin2020mcc}.
They typically aim at building classifiers with domain-invariant features by aligning source and target domains in a feature space.
This requires simultaneous access to data from both the source and target domains.
However, in some fields such as medicine, the period of data accessibility is typically limited, \eg due to strict privacy regulations.
In such scenarios, data from one domain cannot be stored indefinitely to facilitate adaptation to another domain in the future.
Indeed, in real-world settings, often more than one target domain is of concern, necessitating adaptation strategies to multiple target domains that arrive sequentially.
To address such settings, UDA methods are desired that can adapt to a sequence of multiple target domains, without requiring access to data from previously encountered domains.

The above-described problem setting of sequential learning with constrained data access is studied extensively in the field of continual learning (CL)~\cite{Parisi2018ContinualReview,vandeVen2019ThreeLearning,DeLange2019ATasks}.
In this setting, the main challenge becomes catastrophic forgetting~\cite{McCloskey1989,Goodfellow2013AnNetworks}, \ie the stark drop in performance on previously learned tasks when access to previous data is lost.
In this work, we study such a problem setting at the intersection of CL and UDA, which we term \emph{continual UDA}.
While CL has been investigated in several different forms such as class-incremental learning (CIL)~\cite{shin2017dgr,Kemker2017FearNet:Learning,wu2018mergan,ostapenko2019dgm,XiangIncrementalNetworks,yu2020semantic,toldoFusion2022,thandiackal2021genifer} or task-incremental learning (TIL)~\cite{kirkpatrick2017ewc,zenke2017si,nguyen2018vcl,aljundi2018mas}, continual UDA has been explored very little so far.

Earlier works that studied UDA over multiple target domains~\cite{bobu2018iclr,tang2021gradient,taufique2022uclgv} have mainly relied on a replay memory of previously stored samples.
However, this is not in line with the aforementioned continual constraint.
ACE was introduced in~\cite{wu2019ace} based on the idea of transferring the style of target images to labeled source images, thereby creating pseudo-labeled artificial target images.
Although ACE does not employ a replay memory, it requires permanent access to the source domain during the adaptation, which again violates the CL constraint.
IADA~\cite{wulfmeier2018iada} addresses continual UDA by employing \emph{generative feature replay}, a technique that was shown to be effective for CIL~\cite{Kemker2017FearNet:Learning,XiangIncrementalNetworks,Liu2020GenerativeLearning}.
Their main idea is to substitute real source samples with synthetic representations created by a generative adversarial network (GAN)~\cite{Goodfellow2014GANs} to facilitate UDA to a new target domain.
IADA, however, incurs multiple disadvantages:
It is designed for settings with relatively small domain shifts, \eg for gradual adaptation between images of daylight and night-time conditions.
Moreover, IADA only focuses on performance in the final domain of a sequence, thus disregarding the problem of catastrophic forgetting.

\begin{figure}
    \centering
    \includegraphics[width=0.8\linewidth]{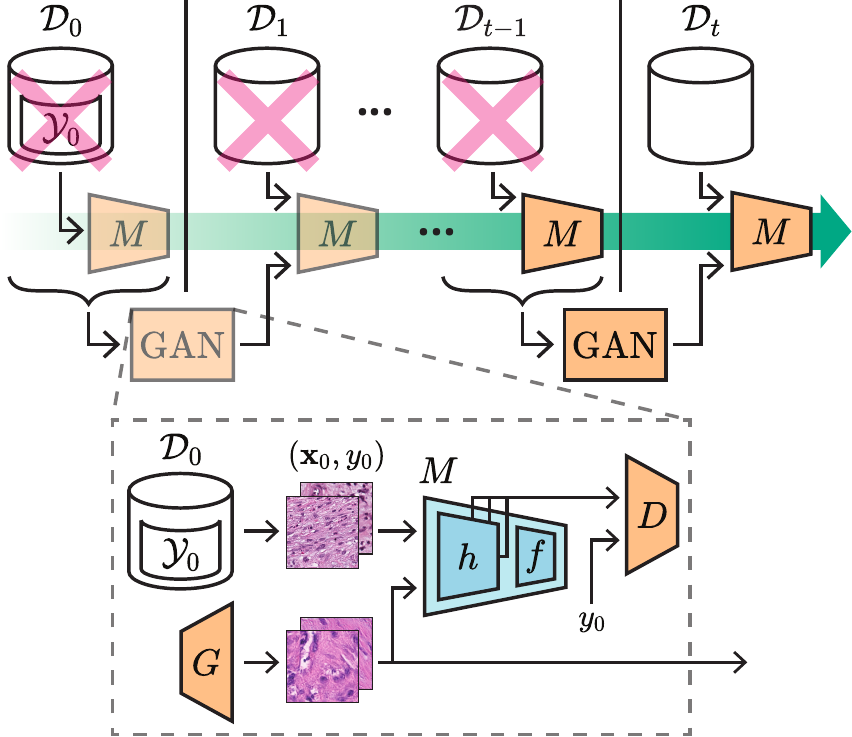}
    \caption{Overview of the continual UDA workflow (top) with a detailed view on our proposed generative feature-driven image replay module with multi-scale feature aggregation (bottom).}
    \label{fig:teaser}
\end{figure}

Given the need for strictly continual methods together with the ability to handle large domain shifts, we herein propose \emph{multi-scale feature alignment} as an extension of generative feature-driven image replay~\cite{thandiackal2021genifer} for continual UDA (\Cref{fig:teaser}).
In contrast to conventional generative replay, the discriminator in this approach does not directly see the images created by the generator, but rather the corresponding image features extracted by the classifier.
Accordingly, all feature extraction layers in the classifier remain trainable, and do not have to be frozen as in conventional generative feature replay.
This additionally allows image-level augmentations on generated samples to further boost classifier performance.
Such feature-driven image replay is herein adapted to UDA, based on inspiration from adversarial discriminative domain adaptation (ADDA)~\cite{tzeng2017adda}.
In ADDA, a domain discriminator is trained to distinguish source and target features, while a feature extractor is trying to confuse the discriminator by aligning source and target features.
Since ADDA's domain discriminator and our GAN discriminator above solve very similar tasks and operate on the feature level, we herein propose to employ the same discriminator for both tasks.
Compared to earlier adversarial UDA approaches, we furthermore enhance the effectiveness of such a discriminator by training it with features aggregated from \emph{multiple} classifier layers, which is empirically shown in this work to improve performance on all domains including the source domain.
Additionally, we incorporate pseudo-labeling and classifier distillation techniques to prevent the feature extractor from learning features that are only useful for domain alignment but not classification.

Contributions in this work include:
(1)~We introduce generative feature-driven image replay for continual UDA and propose a dual-purpose discriminator for GAN training as well as domain adaptation.
(2)~For an optimal trade-off between the performance on previous domains and new target domains, we employ multi-scale aggregation when feeding features to the discriminator.
(3)~We achieve state-of-the-art performance for the task of tissue-type classification, evaluated extensively on domain sequences with three colorectal tissue image datasets and detailed ablation experiments.
(4)~We demonstrate the applicability of our approach to patch-based segmentation on high-resolution tissue images.

\section{Related Work}
\label{sec:rel_work}

\subsection{UDA}
\label{sec:rw_uda}

UDA aims to adapt a model from a labeled source domain to an unlabeled target domain.
For successful UDA, a domain-invariant classifier is required, which makes correct predictions for samples irrespective of their domains.

There exists various approaches that perform UDA by \emph{aligning} the source and target domains.
For instance, PixelDA~\cite{bousmalis2017pixelda} and CyCADA~\cite{hoffman2018cycada} employ GANs conditioned on a source image and a random noise vector to synthesize target domain images.
A classifier is then adapted to the target domain by jointly training on real source and synthetic target images.
Alternative methods focus on domain alignment in a feature space, \eg via adversarial adaptation.
Ganin \etal\cite{ganin2016dann} proposed Domain-Adversarial Neural Network (DANN), which uses a feature-based domain discriminator combined with a gradient reversal layer.
The latter inverts the gradient of the domain discriminator and hence encourages the feature extractor to make the domains indistinguishable, \ie to align them in the feature space.
Such adversarial adaptation is also used in other popular methods such as ADDA~\cite{tzeng2017adda} or CDAN~\cite{long2018cdan}, by employing GAN-inspired losses instead of gradient reversal.

Recently proposed domain adaptation methods exploit custom regularization losses.
For instance, Xu \etal introduced Adaptive Feature Norm (AFN)~\cite{xu2019afn} based on the hypothesis that features with larger norms are more transferable between domains.
In~\cite{jin2020mcc}, a regularization loss called Minimum Class Confusion (MCC) was introduced to reduce the confusion between classes in the target domain, thereby enabling better knowledge transfer from the source to the target.
Adversarial UDA methods have also been effectively applied in CP, \eg for whole-slide image (WSI) classification~\cite{ren2018adversarial}.
Furthermore, self-supervised objectives have been proposed to enable models to learn domain-invariant feature representations~\cite{koohbanani2021selfpath,abbet2022selfrule}.

\subsection{Sequential UDA}
\label{sec:rw_seq_uda}

In contrast to the above-mentioned conventional UDA methods that aim for adaptation to a single target domain, sequential UDA aims at consecutive adaptation to a sequence of target domains.
To that end, Bobu \etal proposed Continuous Unsupervised Adaptation (CUA)~\cite{bobu2018iclr} by employing a replay memory containing a representative set of previously seen samples, so-called exemplars.
Similarly, \cite{tang2021gradient,taufique2022uclgv} also use a replay buffer of previous samples.
Such replay buffers violate continual UDA constraints that prohibit sample storage from earlier domains.
Wu \etal proposed ACE~\cite{wu2019ace}, which tackles UDA in segmentation by pseudo-labeling of artificial images generated via style transfer.
Although ACE does not require an explicit replay memory of exemplars, it still assumes continued access to the source domain, which again violates the CL constraint.
Furthermore, ACE is not suitable for tasks where the domain shift, \ie change of image style, varies largely based on the image class, because no class information is available in unlabeled target datasets.
Such class-dependent domain shift, which is inherent in CP datasets for tissue-type classification, is addressed in this work.

Among methods that fit the CL constraint of no sample storage, IADA~\cite{wulfmeier2018iada} employs \emph{generative feature replay} to replace the missing source samples with synthetic representations learnt and generated via a GAN~\cite{Goodfellow2014GANs}.
However, IADA can only replay the source domain but not any intermediate target domains.
Therefore it is focused on performing well only for the final target domain and not for any intermediate targets.
Consequently, any intermediate targets are ignored and forgotten, thereby reducing the overall performance across domains at the end of the domain sequence.

In contrast to the methods above, our proposed method abides by strictly continual UDA constraints.
Unlike IADA, it does not suffer from catastrophic forgetting of intermediate targets, since our continual GAN training facilitates replay of artificial target samples, aiming to maintain high accuracy on \emph{all} domains seen during a sequence.

\subsection{Generative Replay in CL}
\label{sec:rw_gen_rep}

The idea of generative replay in CL is to replace the real samples that will become inaccessible, with synthetic ones from a generative model, \eg a GAN.
Subsequently, a classifier can be trained jointly on the synthetic samples along with any real samples from a newly introduced domain.

Shin \etal\cite{shin2017dgr} were the first to demonstrate the effectiveness of this approach to replay synthetic images for MNIST~\cite{lecun1998gradient}.
Later, more sophisticated approaches~\cite{wu2018mergan,ostapenko2019dgm,CongGANForgetting} were proposed to tackle problem settings with more complex datasets.
However, training GANs to reproduce high-resolution images is still a major challenge and not a solved problem.
Accordingly, an alternative concept called \emph{generative feature replay}~\cite{Kemker2017FearNet:Learning,XiangIncrementalNetworks,vandeVen2020Brain-inspiredNetworks,Liu2020GenerativeLearning} has gained popularity.
The idea is to replay generated classifier features instead of images.
This is based on the hypothesis that the distribution of features is less complex than the distribution of images and can be better learned and reproduced by a GAN.
Notably, IADA employs generative feature replay (although only to replay the source domain features, not any target features as mentioned earlier).
A common drawback of generative feature replay-based methods~\cite{XiangIncrementalNetworks,vandeVen2020Brain-inspiredNetworks,Kemker2017FearNet:Learning} is their reliance on pretrained (frozen) feature extractors, which prohibits the feature extractor from being adapted to any target data.
This is particularly important in continual UDA as we aim for a classifier with domain-invariant features that are suitable for \emph{all} encountered domains.

In this work, we leverage generative feature-driven image replay, a hybrid form of image and feature replay, first introduced in~\cite{thandiackal2021genifer}.
This method differs from the earlier methods as the generator is trained to synthesize images, while the discriminator only assesses the respective features extracted by the classifier.
Thereby, our method enjoys the benefits of both generative image replay (image-space augmentations to mitigate overfitting) and feature replay (simplified distributions to be learned by the GAN).
In this paper, we leverage this feature-driven construct for an additional advantage in the context of UDA.
We propose to employ a single discriminator both for feature-driven GAN training as well as for feature alignment during classifier adaptation.

\section{Method}
\label{sec:method}

Let $M$ be a model trained with a labeled dataset $\mathcal{D}_0$$=$$\{(\mathbf{x}_0, y_0)^{(i)}\}_{i=1}^{N_0}$ from a source domain, with $N_0$ images $\mathbf{x}_0\in\mathcal{X}_0$ and labels $y_0\in\mathcal{Y}_0$\,.
In continual UDA, the goal is to adapt $M$ to an open sequence of unlabeled datasets
$\mathcal{D}_{1:t:}=\langle\mathcal{X}_1, ..., \mathcal{X}_t, ... \rangle$, 
each sampled from a different target domain.
The terms ``dataset'' and ``domain'' (from which that dataset is sampled) are hereafter used interchangeably.
Since the target datasets do not contain labels, only the labels seen in the source domain can be used for the target domains, \ie no additional labels are introduced after source training.
Further, during the training on the $t$-th target domain, any samples from the source or previously encountered target datasets $\mathcal{D}_{0:t-1}$ are unavailable, and hence $M$ can only access the unlabeled images of the current dataset, \ie $\mathbf{x}_t\in\mathcal{X}_t\equiv\mathcal{D}_t$.

\begin{figure*}
    \centering
    \includegraphics[width=\linewidth]{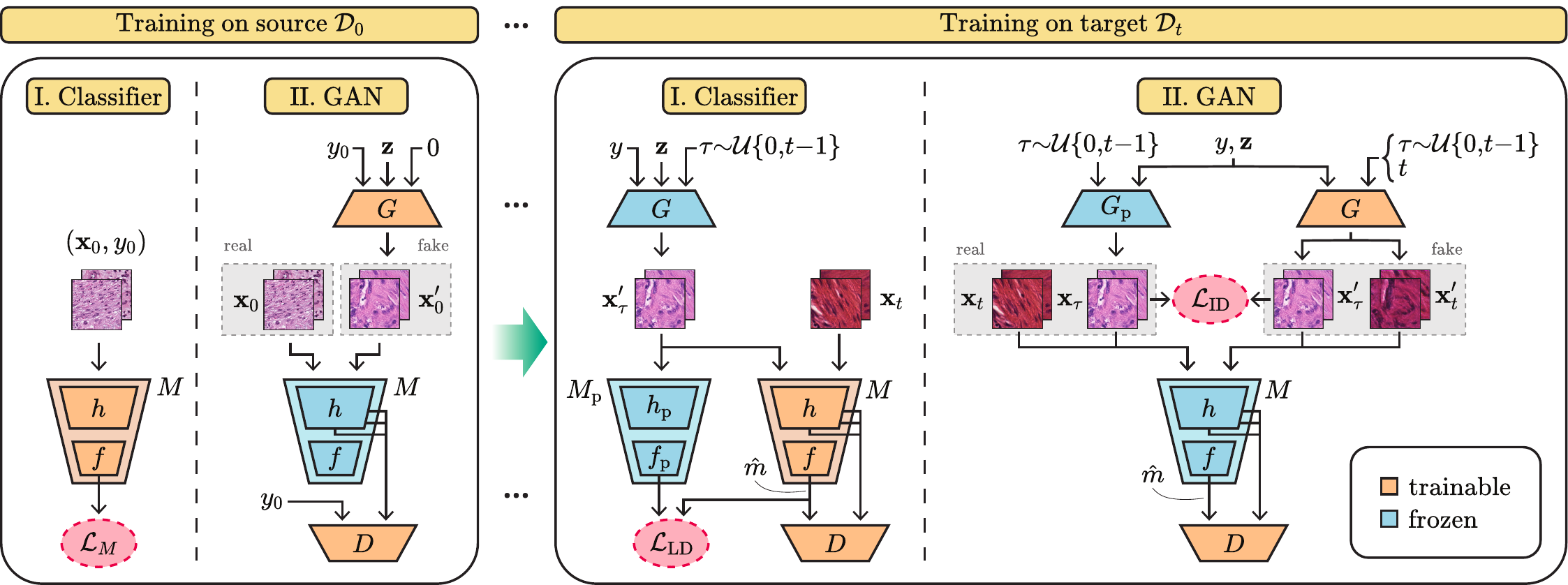}
    \caption{Overview of our method. First, classifier $M$ and then the GAN $(G, D)$ are trained on $\mathcal{D}_0$ (left). For any subsequent domain $\mathcal{D}_{t>0}$, first $M$ and then $(G, D)$ are trained respectively for classifier and GAN adaptation~(right). Domain inputs to $D$ are not shown.}
    \label{fig:overview}
\end{figure*}

The main components of our proposed method are a classifier $M$ and a GAN $(G, D)$, consisting of a generator $G$ and a discriminator $D$.
We train the classifier and GAN in an alternating fashion, by using generative feature-driven image replay~\cite{thandiackal2021genifer} to substitute inaccessible datasets $\mathcal{D}_{0:t-1}$ with artificial images.
\Cref{fig:overview} shows an overview of our proposed method.
First, $M$ and afterwards $(G, D)$ are trained on the labeled source data in a supervised manner.
For any future unlabeled target dataset, first the classifier is adapted to the new domain, followed by the training of $(G, D)$ to reproduce the target dataset.
Subsequently, artificial target samples can be used for replay after the real target data is discarded.

\subsection{Training on the Source Domain}
\label{sec:source_train}

Our classifier $M$$:$$\mathbf{x}$$\mapsto$$f(h(\mathbf{x}))$ consists of a convolutional feature extractor $h$ and a multi-layer perceptron (MLP) $f$ that maps features to output logits for each class.
To obtain output probabilities for each class $l$, we use the softmax function $m^{(l)}(\mathbf{x}) = e^{M^{(l)}(\mathbf{x})} / \sum_{j=1}^{L} e^{M^{(j)}(\mathbf{x})}$.
$M$ is trained on $\mathcal{D}_0$ using a supervised cross-entropy loss as:
\begin{equation}
    \mathop{\mathbb{E}}_{\mathbf{x},y\sim \mathcal{D}_0} \left[ -\sum_{l=1}^{L} 1_{\{y = l\}} \log \left( m^{(l)}(\mathbf{x}) \right) \right].
\end{equation}

Subsequently, we train a conditional GAN consisting of a generator $G$ and a projection discriminator $D$~\cite{Miyato2018CGANsDiscriminator}.
For a random noise vector~$\mathbf{z}$, a class label~$y$, and a domain label~$t$, $G$$:$$(\mathbf{z}, y, t)$$\mapsto$$\mathbf{x}'_t$ learns to synthesize images $\mathbf{x}'_t$ belonging to class~$y$ and domain $t$.
Unless noted otherwise, input noise is uniform, \ie $\mathbf{z}$$\sim$$\mathcal{N}(\mathbf{0}, \mathbf{1})$; and class labels are drawn from the prior label distribution of the source dataset, \ie $y$$\sim$$p_{\mathcal{Y}_0}$.
We define the generator and discriminator losses as:
\begin{IEEEeqnarray}{lcll}
  \mathcal{L}_G &=& \mathop{\mathbb{E}}_{\mathbf{x}'_0\sim p_{G(\mathbf{z}, y, 0)}} &\Big[ a \Big( \!-\!D \big( h(\mathbf{x}'_{0}), y, 0 \big) \Big) \Big]\\
  \mathcal{L}_D &=& \mathop{\mathbb{E}}_{\mathbf{x}'_0\sim p_{G(\mathbf{z}, y, 0)}} &\Big[ a \Big( D \big( h(\mathbf{x}'_{0}), y, 0 \big) \Big) \Big] \nonumber \\
  &&+ \mathop{\mathbb{E}}_{\mathbf{x}, y \sim p_{\mathcal{D}_0}} &\Big[ a \Big( \!-\!D \big( h(\mathbf{x}), y, 0 \big) \Big) \Big] + \lambda_{R_1} R_1 ,
\end{IEEEeqnarray}
where $a(\cdot)$ is the softplus function and $R_1$ is the gradient penalty term $\mathop{\mathbb{E}}_{\mathbf{x} \sim p_{\mathcal{D}_0}} {\left[ \lVert \nabla_{h(\mathbf{x})} D(h(\mathbf{x}), y, 0) \rVert^2_2 \right]}$ from~\cite{Mescheder2018R1}, computed only for real image features and weighted by $\lambda_{R_1}$.

The above has two main differences compared to conventional GANs:
First, we train using feature-driven image replay~\cite{thandiackal2021genifer}, \ie $G$ creates images $\mathbf{x}$ and $D$ sees the corresponding classifier features $h(\mathbf{x})$.
Second, we use \emph{multi-scale feature aggregation} for the discriminator as described below.

\begin{figure}
    \centering
    \includegraphics[height=4cm]{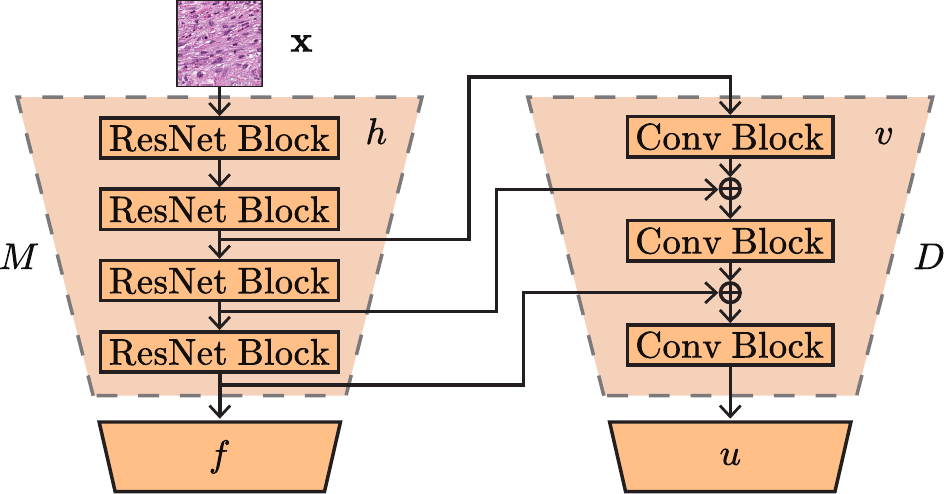}
    \caption{Multi-scale feature aggregation for the discriminator $D$ where $\oplus$ denotes concatenation followed by 1$\times$1 convolution.
    $D$ is shown as split into its feature extraction part $v(\cdot)$ and classification part $u(\cdot)$.}
    \label{fig:mfa}
\end{figure}

{\bf Multi-scale Feature Aggregation (MFA).}
Most adversarial UDA approaches focus on very deep features, \eg from the last layer of feature extractor, which are tightly coupled to classification.
To enable high classification performance across different domains, these deep features need to be domain-invariant.
This is achieved in adversarial UDA by aligning such deep features across domains via a domain discriminator.
On the other hand, for feature-driven image replay it was shown in~\cite{thandiackal2021genifer} that relatively shallower features (\eg at half or three-quarters of the classifier depth) are needed, where the features are still sufficiently general to represent important image variations while also sufficiently specific to filter out information irrelevant to the task at hand.
To consolidate the above aspects of UDA and image replay, we herein propose a single discriminator $D$ used jointly for both tasks, by feeding it with features from \emph{multiple} classifier layers.
Specifically, the features from different classifier layers are ingested in $D$ via concatenation at distinct entry points, as seen in \Cref{fig:mfa}.
We show in ablation experiments~(\Cref{tab:ablations}) that our proposed unified discriminator with MFA is superior to a conventional approach of using independent discriminators for GAN training and UDA, \ie an alternative setting where the discriminators would operate on features from distinct single layers without any aggregation.

\subsection{Training on Target Domains: Classifier Adaptation}
\label{sec:cont_clsf_uda}

For any incoming target dataset $\mathcal{D}_t$, we first adapt the classifier.
As in adversarial UDA, adaptation is achieved herein through a min-max game between the classifier $M$ (mainly its feature extractor $h$) and the domain discriminator $D$; with the goal that $M$ should be able to extract domain-invariant features that enable correct classification across all domains.
As a good initialization, we start training $D$ initialized with the weights from the preceding GAN training, \ie we continue training $D$ from where it was left last time.
For the classifier training, we freeze the previously-trained cumulative classifier $M_{0:t-1}$, where \{0:t-1\} indicates the domains that this classifier can handle, and use its weights to initialize a new trainable classifier $M_t$. 
For simplicity of notation, hereafter the trainable $M_t$ is written as $M$ and the frozen previous $M_{0:t-1}$ as $M_\mathrm{p}$ (with respective $h_\mathrm{p}$ and $f_\mathrm{p}$), as shown in \Cref{fig:overview}.

$D$ aims to predict whether the classifier features are coming from a current or previous domain image, \ie $D$ should output 0 for features $h(\mathbf{x}_t)$ coming from the real images $\mathbf{x}_t$ of the current target, and 1 for features $h(\mathbf{x}'_\tau)$ coming from the synthetic images $\mathbf{x}'_\tau \in \mathcal{D}_{0 : t-1}$ of any previous domain.
Note that $D$ does not consider any domain label as input, as the task would become trivial otherwise.
However, since images belonging to different class labels may naturally look very different, $D$ should judge the images in the context of their class label, so $D$ needs a respective class label as input.
Therefore, we compute pseudo-labels using our classifier, \ie $\hat{m}(\mathbf{x}_t)$$=$$\arg\max_l m^{(l)}(\mathbf{x}_t)$ and $\hat{m}(\mathbf{x}'_\tau)$$=$$\arg\max_l m^{(l)}(\mathbf{x}'_\tau)$.
The objective for $D$ can then be defined as a binary cross-entropy loss with added $R_1$ regularization (as defined in \Cref{sec:source_train}):
\begin{IEEEeqnarray}{lcl}
\label{eq:d_uda_loss}
  \mathcal{L}_D &=& \mathop{\mathbb{E}}_{\mathbf{x}_t\sim p_{\mathcal{X}_t}} \Big[ -\log \Big(1- D\big(h(\mathbf{x}_t), \hat{m}(\mathbf{x}_t) \big) \Big) \Big] \nonumber \\
  &&+ \!\!\!\!\mathop{\mathbb{E}}_{\substack{\tau \sim \mathcal{U}\{0, t-1\} \\ \mathbf{x}'_\tau \sim p_{G(\mathbf{z}, y, \tau)}}} \!\!\Big[\!-\log \!\Big( D\big(h(\mathbf{x}'_\tau), \hat{m}(\mathbf{x}'_\tau)\big) \Big) \Big] \!+ \!\lambda_{R_1} \!R_1 .
\end{IEEEeqnarray}

Meanwhile, the classifier $M$ aims to confuse $D$ so that $D$ cannot differentiate features from different domains, thus achieving domain invariance.
This is realized effectively with a binary cross-entropy.
To prevent $M$ from forgetting its task on any previous domains, we want to encourage the currently trained classifier $M$ to make the same predictions as its frozen copy $M_\mathrm{p}$, for any generated synthetic images belonging to previous domains.
To that end, we utilize an additional logit distillation loss~\cite{Hinton2015DistillingNetwork}  $\mathcal{L}_\mathrm{LD}$, weighted by $\lambda_\mathrm{LD}$.
These lead to the classifier loss defined as:
\begin{IEEEeqnarray}{lcl}
\label{eq:m_uda_loss}
  \mathcal{L}_M &=& \mathop{\mathbb{E}}_{\mathbf{x}_t\sim p_{\mathcal{X}_t}} \Big[ -\log \Big( D \big(h(\mathbf{x}_t), \hat{m}(\mathbf{x}_t) \big) \Big) \Big] \nonumber \\
  &&+\lambda_\mathrm{LD} \!\!\underbrace{\mathop{\mathbb{E}}_{\substack{\tau \sim \mathcal{U}\{0, t-1\} \\ \mathbf{x}'_\tau \sim p_{G(\mathbf{z}, y, \tau)}}} \Big[ \big\lVert f_\mathrm{p} \big( h_\mathrm{p}(\mathbf{x}'_\tau) \big) - f \big( h(\mathbf{x}'_\tau) \big) \big\rVert_1 \Big]}_{\mathcal{L}_\mathrm{LD}}\!.
\end{IEEEeqnarray}
Note that $\mathcal{L}_\mathrm{LD}$ also acts as a regularizer to ensure that $h$ and $f$ do not diverge from each other, \ie it prevents $h$ from learning to extract features that are only useful for domain alignment but not for classification.

\subsection{Training on Target Domains: GAN Adaptation}
\label{sec:cont_gan}

Subsequent to classifier adaptation, the generator $G$ is adapted to also reproduce images from the target domain $\mathcal{D}_t$ so that the real images do not have to be retained for future adaptation.
$D$ and $G$ are initialized with weights from the previously trained GAN, and a frozen copy $G_\mathrm{p}$ of the previous generator is retained temporarily only during the training on the current domain $\mathcal{D}_t$.
Since previous real data is not accessible anymore, we have to mitigate forgetting in $G$, \ie it should remember how to create images of all previous domains $\mathcal{D}_{0:t-1}$.
To this end, the generator is trained with losses that integrate not only the objective of learning to synthesize new data, but also to retain the ability of creating previous data:
\begin{IEEEeqnarray}{lcl}
\label{eq:g_cont_loss}
  \mathcal{L}_G &=& \mathop{\mathbb{E}}_{\substack{\tau \sim \mathcal{U}\{0, t\}\\\mathbf{x}'_\tau\sim p_{G(\mathbf{z}, y, \tau)}}} \Big[ a \Big(-D\big(h(\mathbf{x}'_\tau), \hat{m}(\mathbf{x}'_\tau), \tau \big) \Big) \Big] \nonumber \\
  &&+ \lambda_\mathrm{ID} \underbrace{\mathop{\mathbb{E}}_{\tau \sim \mathcal{U}\{0, t-1\}} \Big[ \big\lVert G(\mathbf{z}, y, \tau) - G_\mathrm{p}(\mathbf{z}, y, \tau) \big\rVert_1 \Big]}_{\mathcal{L}_\mathrm{ID}} ,
\end{IEEEeqnarray}
where the first term is adversarial, \ie driven by the feedback of the discriminator, and the second term is an image distillation loss $\mathcal{L}_\mathrm{ID}$.
The latter distillation drives $G$ to generate the same images as its previous copy, when given the same random input noise, class label, and domain label from any previous domain.

The discriminator is meanwhile trained to distinguish real and fake classifier features coming from images of all previous domains $\mathcal{D}_{0:t-1}$ as well as the current domain $\mathcal{D}_t$:
\begin{IEEEeqnarray}{lcl}
\label{eq:d_cont_loss}
  \mathcal{L}_D &=& \mathop{\mathbb{E}}_{\substack{\tau \sim \mathcal{U}\{0, t\}\\\mathbf{x}'_\tau\sim p_{G(\mathbf{z}, y, \tau)}}} \Big[ a \Big(D\big(h(\mathbf{x}'_\tau), \hat{m}(\mathbf{x}'_\tau), \tau \big) \Big) \Big] \nonumber \\
  &&+ \mathop{\mathbb{E}}_{\substack{\tau \sim \mathcal{U}\{0, t-1\}\\\mathbf{x}_\tau\sim p_{G_\mathrm{p}(\mathbf{z}, y, \tau)}}} \Big[ a \Big(-D\big(h(\mathbf{x}_\tau), \hat{m}(\mathbf{x}_\tau), \tau \big) \Big) \Big] \nonumber \\
  &&+ \mathop{\mathbb{E}}_{\mathbf{x}_t \sim p_{\mathcal{X}_t}} \Big[ a \Big(-D\big(h(\mathbf{x}_t), \hat{m}(\mathbf{x}_t), t \big) \Big) \Big] + \!\lambda_{R_1} \!R_1 .
  \label{eqn:gan_loss}
\end{IEEEeqnarray}
After the GAN training completes, the real images from $\mathcal{D}_t$ can be discarded.
At the arrival of a future domain/dataset $\mathcal{D}_{t+1}$, these discarded images can then be replaced with the artificially generated images $\mathbf{x}'_t$$=$$G(\mathbf{z}, y, t)$ to replay during classifier adaptation.

\section{Experiments}
\label{sec:exp}

\subsection{Datasets}
\label{sec:data}
We herein consider image patches extracted from H\&E-stained (Hematoxylin \& Eosin) WSIs of colorectal biopsies.
In particular, we utilize three datasets: \mbox{K-16}~\cite{kather2016data,kather2016}; \mbox{K-19}~\cite{kather2019data,kather2019}; and \mbox{CRC-TP}~\cite{javed2020crctp}.
These datasets were also used for benchmarking in previous work on UDA~\cite{abbet2022selfrule}.
All datasets are digitized at 20$\times$ magnification.
\mbox{K-19} contains 107\,180 patches of size 224$\times$224 pixels (.5\,$\mu$m/pixel) from 9 classes.
We stratify and randomly split the data into 85-15\% for training-validation, and use the official test set containing 7180 samples. 
\mbox{K-16} has 5000 patches of size 150$\times$150 pixels (.495\,$\mu$m/pixel) from 8 classes. 
We employ a random stratified split of 70-15-15\% into training-validation-test sets. 
\mbox{CRC-TP} consists of 280\,000 patches with a size of 150$\times$150 pixels (.495\,$\mu$m/pixel), but only 7 classes.
We use a 85-15\% split into training-validation sets, and use the official test set containing 69\,000 samples.
For each dataset, the test set does not have any patients in common with the training or validation sets.
As the datasets have different patch sizes, all patches are center-cropped to 128$\times$128 pixels.
Following~\cite{abbet2022selfrule}, we rearrange and align the heterogeneous classes from the different datasets into a common set of 7 medically-relevant patch classes: tumor epithelium (TUM), stroma (STR), lymphocytes (LYM), debris (DEB), normal mucosal glands (NORM), adipose (ADI), background (BACK).

The CL domain sequences to use in the evaluation are herein defined based on the following observations:
First, \mbox{CRC-TP} does not contain the classes ADI and BACK present in the other domains. 
Thus, selecting \mbox{CRC-TP} as the source domain would amount to an open-set UDA setting, beyond the focus of this work.
Nonetheless, note that having \mbox{CRC-TP} as a target dataset puts us in a partial-set UDA setting, which requires mitigation techniques against forgetting of the missing classes.
Second, K-16 is a relatively small dataset.
As such, it would unlikely be used as source in practice, where a reasonable effort can be expected to curate the initial source model. 
Therefore, \mbox{K-19} is chosen as the source domain, and for the target domains we consider both possible orders. 
Accordingly, naming \mbox{K-19} as $\mathcal{D}_0$, \mbox{K-16} as $\mathcal{D}_1$, and \mbox{CRC-TP} as $\mathcal{D}_2$, we experimentally study the two continual domain sequences: $\mathcal{D}_0$$\rightarrow$$\mathcal{D}_1$$\rightarrow$$\mathcal{D}_2$ and $\mathcal{D}_0$$\rightarrow$$\mathcal{D}_2$$\rightarrow$$\mathcal{D}_1$.

\subsection{Implementation Details}
\label{sec:implementation}

For the classifier $M$, we use ResNet-18~\cite{He2016DeepRecognition} with pretrained weights~\cite{russakovsky2015imagenet} as the feature extractor $h$, and average pooling followed by two linear layers as classifier head $f$.
For each method and adaptation step, we trained for at most 200 epochs, stopping before if an early stopping criterion is satisfied:
On $\mathcal{D}_0$, the early stopping criterion is based on the classification loss on the validation set.
On the unlabeled target domains, we employ the negative of $\mathrm{InfoMax}$~\cite{Shi2012infomax} as a surrogate loss, \ie $\frac{1}{N} \sum_{i=1}^{N} \mathbf{H}(p_i) - \mathbf{H}(\frac{1}{N} \sum_{i=1}^{N} p_i)$, 
where $\mathbf{H}(\cdot)$ is the entropy function and $p_i$ are the predicted class probabilities for the $i$-th sample in the validation set.
We use the RAdam~\cite{Liu2020OnBeyond} optimizer with a cosine scheduler and a weight decay of 0.01.
We augment the training images via random cropping, horizontal and vertical flips, and random affine transformations.
For all methods, we employed a batch size of 16.

For generative replay, we use a conditional GAN with projection discriminator~\cite{Miyato2018CGANsDiscriminator}.
It has seven convolutional layers and a mini-batch discrimination layer~\cite{karras2017progressive} for better sample diversity.
Our generator has two linear layers followed by nine style-convolution layers~\cite{Karras2020AnalyzingStyleGAN}.
The GAN was optimized using Adam~\cite{Kingma2014Adam} with respective learning rates 0.0002 and 0.0003 for $G$ and $D$.
We apply exponential moving averaging on generator weights~\cite{Yazc2019TheTraining} to avoid ending the training in an unstable state.
The GAN was trained for 250\,000 steps on the source and 150\,000 steps on the first target domain.
For our method, we searched the best hyperparameters on a grid of values comprising $\{0.5, 1, 2\}$ for logit distillation $\lambda_\mathrm{LD}$~\eqref{eq:m_uda_loss}, $\{0.5, 1, 1.5\}$ for image distillation $\lambda_\mathrm{ID}$~\eqref{eq:g_cont_loss}, and $\{1, 2, 5, 10\}$ for regularization weight $\lambda_{R_1}$, which is used for adapting both the classifier~\eqref{eq:d_uda_loss} and the GAN~\eqref{eq:d_cont_loss}.
All models were implemented in PyTorch~\cite{Paszke2019PyTorch:Library} and trained on an NVIDIA A100 GPU.

\subsection{Comparisons with the State of the Art}
\label{sec:baselines}

We compare our method against DANN~\cite{ganin2016dann}, ADDA~\cite{tzeng2017adda}, CDAN~\cite{long2018cdan}, AFN~\cite{xu2019afn}, and MCC~\cite{jin2020mcc}, as they are either seminal or state-of-the-art methods in the field of adversarial UDA.
Since these methods are not directly applicable to our continual UDA setting, we equip them with generative replay of source images to facilitate a fair comparison.
To this end, we employ almost the same GAN architecture as in our method, with the difference that the discriminator operates directly on images and not features.
In addition, we report the lower bound (LB), which is a model only trained on the source without any adaptation; the single-domain upper bound (SD-UB), where a supervised model is trained and tested on the same target domain; and the multi-domain upper bound (MD-UB), which represents a supervised model trained and tested jointly on all domains.
Note that SD-UB and MD-UB can be considered as hypothetical ``oracles'', since the target domain labels would not be available in the actual considered UDA setting.

\subsection{Results}
\label{sec:results}
\subsubsection{Classification Performance}
\label{sec:clsf_perf}
As mentioned above, we consider two continual domain sequences.
The classification performance is measured using weighted F1 score, attained \emph{at the end} of a sequence.
We present the domain-wise F1 scores as well as the overall F1 score averaged across all domains.
The domain-wise scores on the target domains demonstrate the adaptation capability of the model.
To evaluate and compare different methods, we report the overall F1 score of each approach over all considered domains.

The classification results of our approach and the competing methods are shown in \Cref{tab:clsf_overall}(right).
\begin{table*}
    \setlength{\tabcolsep}{2pt}
    \notsotiny
    \caption{Test F1 scores (\%) at the end of the continual sequences $\mathcal{D}_0$$\rightarrow$$\mathcal{D}_1$$\rightarrow$$\mathcal{D}_2$ (top half) and $\mathcal{D}_0$$\rightarrow$$\mathcal{D}_2$$\rightarrow$$\mathcal{D}_1$ (bottom half).
    Results are shown as mean $\pm$ standard deviation (std-dev) over 5 random initializations, where the best and the second-best results per column (excluding LB and UB) are in \textbf{bold} and \underline{underlined}, respectively. 
    Per-class F1 scores (aggregated over all domains) are shown on the left, per-domain and overall F1 scores are on the right.}
    \label{tab:clsf_overall}
    \centering
    \begin{tabular}{llcccccccccccc}
        \toprule
        \multirow{2}[1]{*}{Training} & \multirow{2}[1]{*}{Method} & \multicolumn{3}{c}{F1 per Domain} & \multirow{2}[1]{*}{\bf F1 Overall} & \multicolumn{7}{c}{F1 per Class} \\
        \cmidrule(lr){3-5} \cmidrule(lr){7-13}
        && $\mathcal{D}_0$ & $\mathcal{D}_1$ & $\mathcal{D}_2$ && TUM & STR & LYM & DEB & NORM & ADI & BACK\\
        \midrule
        $\mathcal{D}_0$ & LB & \cellcolor{grady-med}{$99.5$} & \cellcolor{grady-med}{$33.4$} & \cellcolor{grady-med}{$31.9$} & \cellcolor{grady-dark}{$54.9$} & $8.0$ & $26.2$ & $0.0$ & $23.5$ & $29.2$ & $69.2$ & $0.0$ \\
        \midrule
        \multirow{6}{*}{$\mathcal{D}_0$$\rightarrow$$\mathcal{D}_1$$\rightarrow $$\mathcal{D}_2$}
         & DANN~\cite{ganin2016dann} & \cellcolor{grady-med}{$86.4\pm0.5$} & \cellcolor{grady-med}{$66.2\pm2.0$} & \cellcolor{grady-med}{$62.8\pm1.2$} & \cellcolor{grady-dark}{$71.8\pm0.9$} & $74.4\pm2.3$ & $68.6\pm2.0$ & $53.2\pm1.8$ & $39.6\pm2.9$ & $48.8\pm1.9$ & $77.6\pm7.1$ & $82.6\pm4.3$\\
         & ADDA~\cite{tzeng2017adda} & \cellcolor{grady-med}{$81.1\pm1.0$} & \cellcolor{grady-med}{$50.9\pm2.5$} & \cellcolor{grady-med}{$62.9\pm1.2$} & \cellcolor{grady-dark}{$65.0\pm1.2$} & $72.4\pm1.6$ & $38.8\pm2.5$ & $\mathbf{64.8\pm1.9}$ & $46.0\pm2.2$ & $\underline{65.7\pm1.7}$ & $62.9\pm2.7$ & $29.3\pm12.2$\\
         & CDAN~\cite{long2018cdan} & \cellcolor{grady-med}{$85.7\pm0.5$} & \cellcolor{grady-med}{$70.9\pm1.0$} & \cellcolor{grady-med}{$63.5\pm0.6$} & \cellcolor{grady-dark}{$73.3\pm0.5$} & $73.0\pm3.6$ & $\underline{75.8\pm1.3}$ & $53.0\pm1.5$ & $47.1\pm2.2$ & $48.2\pm1.8$ & $\underline{81.8\pm3.0}$ & $\underline{87.2\pm1.8}$\\
         & AFN~\cite{xu2019afn} & \cellcolor{grady-med}{$86.7\pm0.6$} & \cellcolor{grady-med}{$67.6\pm1.8$} & \cellcolor{grady-med}{$59.0\pm1.9$} & \cellcolor{grady-dark}{$71.1\pm1.1$} & $74.2\pm3.0$ & $64.6\pm4.2$ & $58.0\pm2.8$ & $38.6\pm2.7$ & $56.6\pm3.3$ & $78.3\pm4.7$ & $79.3\pm2.8$\\
         & MCC~\cite{jin2020mcc} & \cellcolor{grady-med}{$\underline{87.0\pm0.8}$} & \cellcolor{grady-med}{$\underline{72.5\pm2.3}$} & \cellcolor{grady-med}{$\underline{66.0\pm2.4}$} & \cellcolor{grady-dark}{$\underline{75.2\pm1.4}$} & $\underline{80.6\pm2.1}$ & $75.0\pm2.5$ & $52.3\pm4.1$ & $\underline{50.7\pm3.1}$ & $55.9\pm2.5$ & $80.7\pm7.5$ & $85.8\pm3.1$\\
         & Ours & \cellcolor{grady-med}{$\mathbf{87.5\pm0.5}$} & \cellcolor{grady-med}{$\mathbf{86.8\pm0.7}$} & \cellcolor{grady-med}{$\mathbf{71.9\pm0.8}$} & \cellcolor{grady-dark}{$\mathbf{82.1\pm0.5}$} & $\mathbf{83.6\pm1.1}$ & $\mathbf{80.3\pm1.9}$ & $\underline{62.6\pm1.1}$ & $\mathbf{61.6\pm2.1}$ & $\mathbf{81.5\pm1.2}$ & $\mathbf{94.9\pm0.8}$ & $\mathbf{95.1\pm0.4}$\\
        \midrule
        \multirow{6}{*}{$\mathcal{D}_0$$\rightarrow$$\mathcal{D}_2$$\rightarrow$$\mathcal{D}_1$}
         & DANN~\cite{ganin2016dann} & \cellcolor{grady-med}{$88.1\pm0.5$} & \cellcolor{grady-med}{$78.6\pm1.4$} & \cellcolor{grady-med}{$53.7\pm2.9$} & \cellcolor{grady-dark}{$73.5\pm1.2$} & $65.5\pm5.5$ & $78.4\pm1.4$ & $54.9\pm3.4$ & $47.4\pm2.4$ & $45.7\pm3.2$ & $88.4\pm3.2$ & $92.9\pm1.2$\\
         & ADDA~\cite{tzeng2017adda} & \cellcolor{grady-med}{$83.6\pm2.0$} & \cellcolor{grady-med}{$78.3\pm 1.9$} & \cellcolor{grady-med}{$\underline{57.4\pm1.3}$} & \cellcolor{grady-dark}{$73.1\pm1.3$} & $\underline{72.0\pm2.6}$ & $67.9\pm3.1$ & $49.8\pm2.9$ & $\underline{53.2\pm2.6}$ & $\underline{66.4\pm2.3}$ & $87.1\pm1.3$ & $88.0\pm1.1$\\
         & CDAN~\cite{long2018cdan} & \cellcolor{grady-med}{$\underline{88.2\pm0.4}$} & \cellcolor{grady-med}{$80.0\pm1.0$} & \cellcolor{grady-med}{$57.2\pm2.3$} & \cellcolor{grady-dark}{$\underline{75.2\pm0.9}$} & $68.9\pm4.4$ & $\mathbf{80.3\pm1.0}$ & $54.9\pm3.0$ & $50.0\pm1.4$ & $52.5\pm1.6$ & $88.6\pm5.1$ & $93.6\pm2.3$\\
         & AFN~\cite{xu2019afn} & \cellcolor{grady-med}{$87.9\pm0.5$} & \cellcolor{grady-med}{$79.0\pm1.6$} & \cellcolor{grady-med}{$49.6\pm2.8$} & \cellcolor{grady-dark}{$72.2\pm1.2$} & $66.4\pm5.3$ & $69.8\pm4.1$ & $58.2\pm5.4$ & $42.2\pm2.6$ & $56.3\pm5.2$ & $80.9\pm5.7$ & $\underline{94.4\pm1.2}$ \\
         & MCC~\cite{jin2020mcc} & \cellcolor{grady-med}{$87.9\pm0.9$} & \cellcolor{grady-med}{$\underline{81.0\pm2.1}$} & \cellcolor{grady-med}{$53.6\pm2.7$} & \cellcolor{grady-dark}{$74.2\pm1.4$} & $61.7\pm4.8$ & $78.2\pm1.5$ & $\underline{58.5\pm4.2}$ & $48.8\pm2.2$ & $53.7\pm6.5$ & $\mathbf{91.9}\pm1.6$ & $\mathbf{95.0}\pm0.4$\\
         & Ours & \cellcolor{grady-med}{$\mathbf{88.8\pm0.4}$} & \cellcolor{grady-med}{$\mathbf{82.2\pm0.7}$} & \cellcolor{grady-med}{$\mathbf{70.6\pm0.7}$} & \cellcolor{grady-dark}{$\mathbf{80.5\pm0.5}$} & $\mathbf{79.5\pm1.1}$ & $\underline{79.8\pm1.5}$ & $\mathbf{64.7\pm1.4}$ & $\mathbf{54.9\pm1.5}$ & $\mathbf{73.1\pm1.5}$ & $\underline{90.5\pm2.1}$ & $92.8\pm1.4$\\
        \midrule
        $\mathcal{D}_1$ & SD-UB  & \cellcolor{grady-med}{$-$} & \cellcolor{grady-med}{$98.2$} & \cellcolor{grady-med}{$-$} & \cellcolor{grady-dark}{$-$} & $97.8$ & $96.8$ & $96.9$ & $98.9$ & $98.9$ & $98.9$ & $98.9$ \\
        $\mathcal{D}_2$ & SD-UB & \cellcolor{grady-med}{$-$} & \cellcolor{grady-med}{$-$} & \cellcolor{grady-med}{$90.3$} & \cellcolor{grady-dark}{$-$} & $93.6$ & $94.2$ & $82.8$ & $88.1$ & $81.1$ & $-$ & $-$\\
        $\mathcal{D}_0$$+$$\mathcal{D}_1$$+$$\mathcal{D}_2$ & MD-UB & \cellcolor{grady-med}{$97.4$} & \cellcolor{grady-med}{$97.3$} & \cellcolor{grady-med}{$88.3$} & \cellcolor{grady-dark}{$94.3$} & $95.2$ & $95.6$ & $90.3$ & $94.4$ & $90.8$ & $98.0$ & $98.3$\\
        \bottomrule
    \end{tabular}
\end{table*}
Our method is seen to achieve the highest F1 score across all domains, consistently for both domain sequences.
For instance, on the first sequence $\mathcal{D}_0$$\rightarrow$$\mathcal{D}_1$$\rightarrow$$\mathcal{D}_2$, ours outperforms the second-best MCC by 14.3\,pp (percentage points) on $\mathcal{D}_1$ and 5.9\,pp on $\mathcal{D}_2$, while maintaining 0.5\,pp better F1 on $\mathcal{D}_0$.
Together, this results in an improvement of 6.9\,pp in mean overall F1, \ie an increment of 33\% compared to MCC's 20.3\,pp difference from LB.
Moreover, ours brings the state of the art 36\% closer to the upper bound, \ie MD-UB, compared to the best existing method, MCC.
Similar trends are also observed for the second sequence studied.
We provide detailed class-wise F1 scores (averaged across the tested domains) on the left half of \Cref{tab:clsf_overall}.
The per-class results show that in the first domain sequence, our method performs best for all classes except LYM, where it still achieves the second-best result.
Similarly, in the second sequence, ours obtains either the best or the second-best F1 scores, with the exception of the BACK class (where still a relatively high score of 92.8\% is achieved).

\begin{figure}
     \centering
     \begin{subfigure}{0.48\linewidth}
         \centering
         \includegraphics[width=0.95\linewidth]{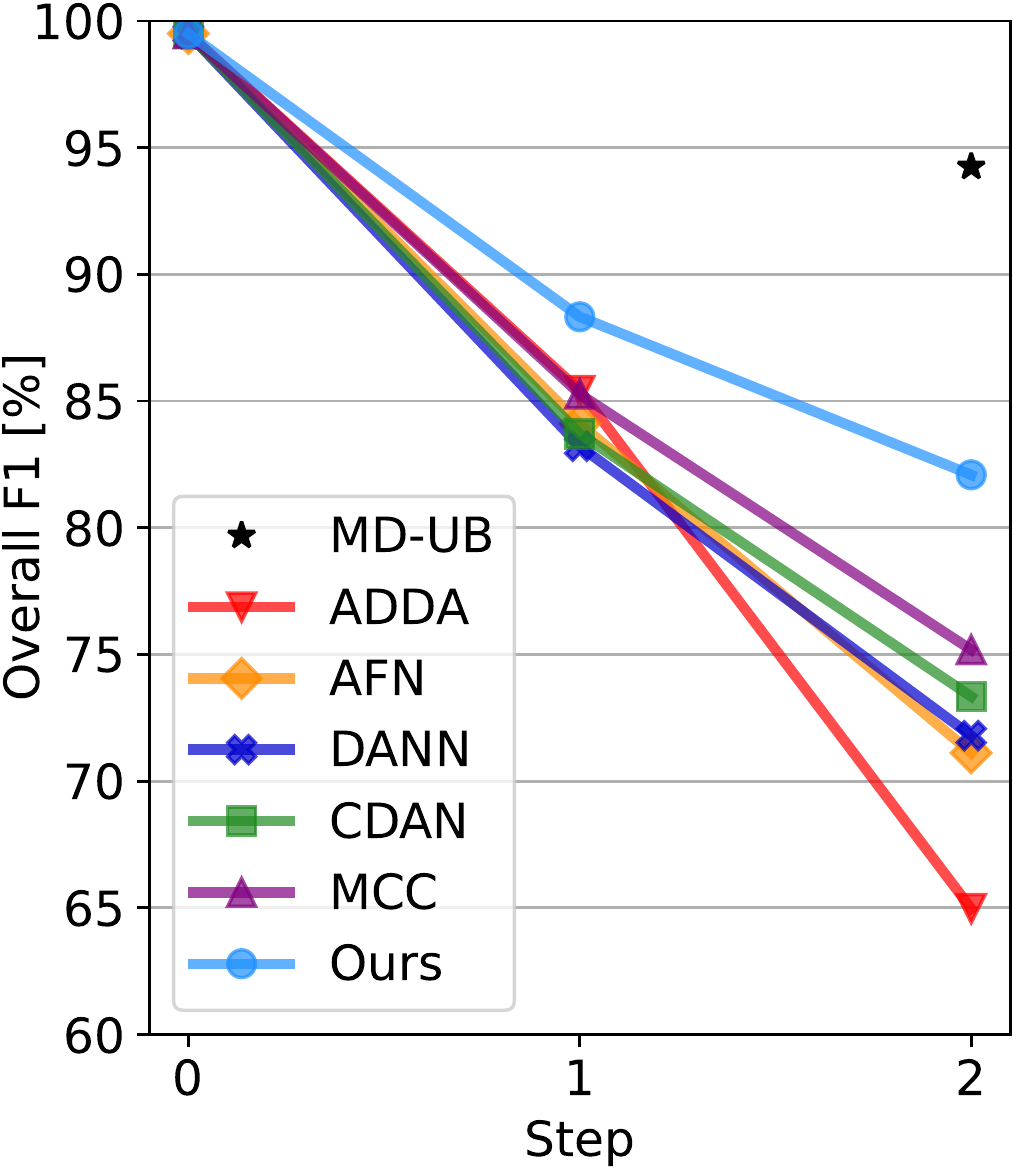}
         \caption{$\mathcal{D}_0 \rightarrow \mathcal{D}_1 \rightarrow \mathcal{D}_2$}
         \label{fig:avg_f1_evol_123}
     \end{subfigure}
     \begin{subfigure}{0.48\linewidth}
         \centering
         \includegraphics[width=0.95\linewidth]{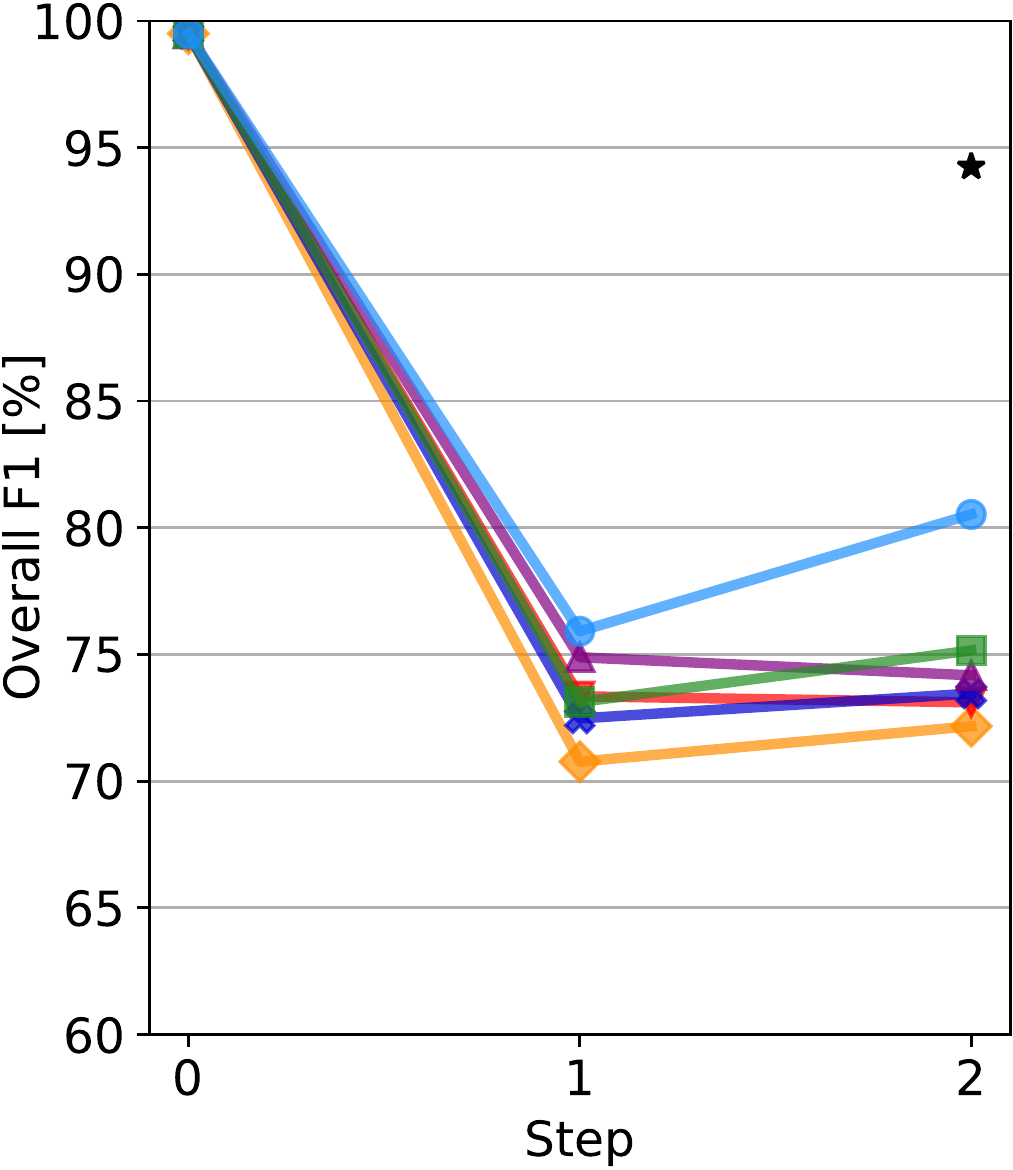}
         \caption{$\mathcal{D}_0 \rightarrow \mathcal{D}_2 \rightarrow \mathcal{D}_1$}
         \label{fig:avg_f1_evol_132}
     \end{subfigure}
     \caption{Evolution of the mean overall F1 score after each adaptation step, shown for two domain sequences.}
     \label{fig:avg_f1_evol}
\end{figure}

We show in \Cref{fig:avg_f1_evol} the evolution of the mean overall F1 score after each adaptation step.
Interestingly, the trends of evolution differ between the sequences.
The overall F1 score decreases monotonically in the first sequence~(\Cref{fig:avg_f1_evol_123}), corroborating a natural expectation.
However, during the second sequence~(\Cref{fig:avg_f1_evol_132}), the overall F1 score stays almost constant or even increases during the last step, \ie with the final addition of domain $\mathcal{D}_1$.
This can be attributed to $\mathcal{D}_2$ being a relatively more difficult domain for classification compared to the other two, as also evident from the MD-UB comparison in \Cref{tab:clsf_overall}.
Notwithstanding, in both sequences, our method obtains the highest overall F1 score at each adaptation step.
Furthermore, its gap to the second-best method increases significantly when going from the first to the second target.

\subsubsection{Analysis of Feature Alignment}
\label{sec:feat_alignment}

\begin{figure*}
     \centering
     \includegraphics[width=\linewidth]{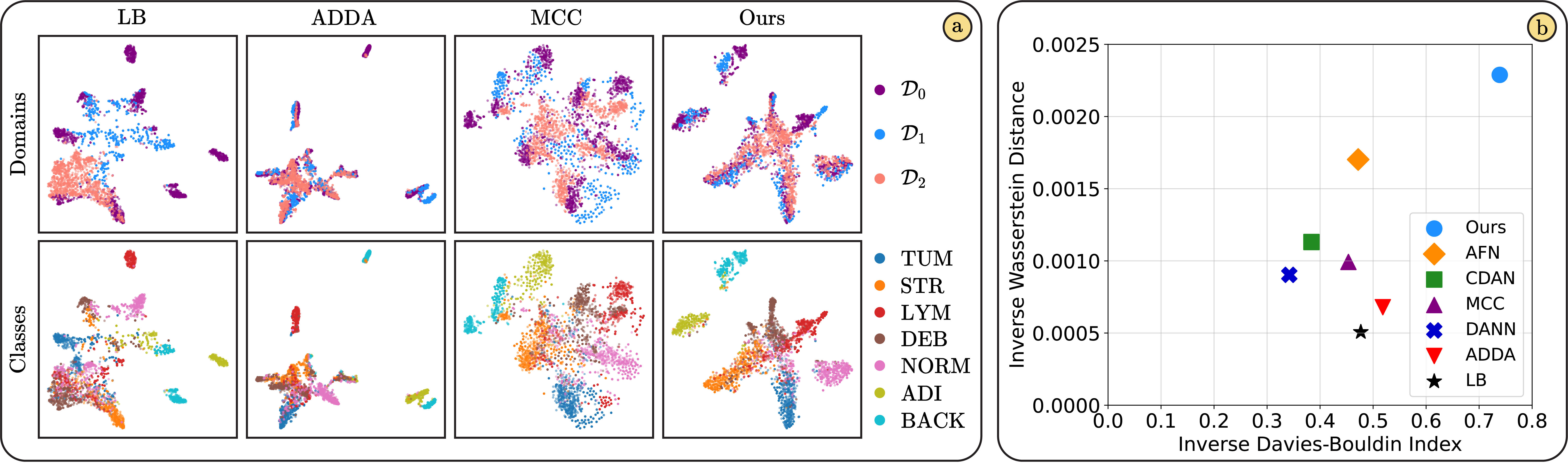}
     \caption{Feature alignment and clustering for different UDA methods. (a) t-SNE feature representations of 300 test images, color-coded in the top row to distinguish different domains and in the bottom row to distinguish different classes.
     Note that $D_2$ does not contain two labels, represented by the two isolated upper-left clusters from our method.
     (b) Domain alignment is quantified via the inverse Wasserstein distance (y-axis) and class clustering via the inverse Davies-Bouldin index (x-axis).}
     \label{fig:feat_align}
\end{figure*}

To get a better understanding of domain alignment in the feature space, we examine the distribution of features at the end of the continual domain sequence $\mathcal{D}_0$$\rightarrow$$\mathcal{D}_1$$\rightarrow$$\mathcal{D}_2$.
In \Cref{fig:feat_align}(a), we present t-SNE embeddings extracted using different UDA methods.
The plots show the features from a fixed random subset of 300 test images for each class.
One would expect the embeddings from an ideal method to differentiate each class into clusters (to achieve the classification task) and to distribute across domains rather uniformly (for domain invariance).
One can see that if the classifier is only trained on the source domain (LB), the features extracted from different domains are not well aligned.
ADDA and MCC seem to achieve better domain alignment.
ADDA class clusters appear to be more compact than in the case of MCC, but both these methods exhibit overlaps between different classes, potentially affecting their classification performance.
Our method demonstrates both good domain alignment and separable class clusters.
This can also be inferred from \Cref{fig:feat_align}(b) where we quantify each method in terms of its domain alignment (y-axis) and class clustering quality (x-axis).
The former is computed using the inverse Wasserstein distance $1/d_\mathrm{W}$ and the latter using the inverse Davies-Bouldin index $1/I_\mathrm{DB}$~\cite{davies1979cluster} for the features extracted from the above-mentioned test images.
The closer a method is placed to the upper right corner, the better it is in terms of achieving both objectives of domain alignment and class clustering.
Our method is seen to attain the highest values for both metrics.

\subsubsection{Qualitative Analysis of Generated Images}
\label{sec:qualitative}

\begin{figure}
    \centering
    \includegraphics[width=\linewidth]{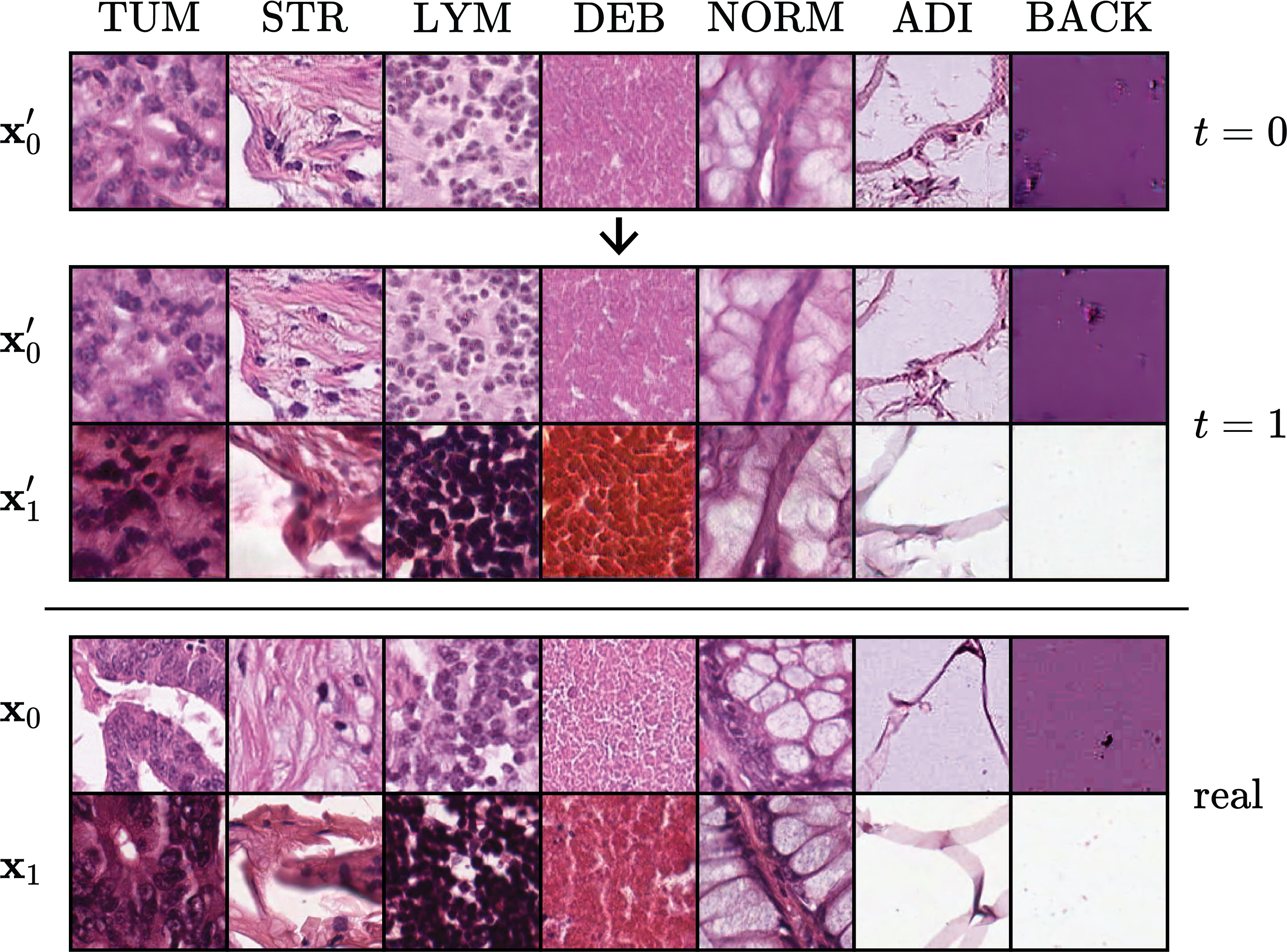}
    \caption{Example images for each class sampled from our generator (top) after steps $t=0,1$ and real images sampled from $\mathcal{D}_0$ and $\mathcal{D}_1$ (bottom).}
    \label{fig:gen_images}
\end{figure}

In \Cref{fig:gen_images}, we present example images, real ones as well as images sampled from our generator.
For a domain sequence $\mathcal{D}_0$$\rightarrow$$\mathcal{D}_1$, we present images generated after the initial source training on $\mathcal{D}_0$ as well as after adaptation to target domain $\mathcal{D}_1$.
Note that after such adaptation, the generator is expected to reproduce images from both of these domains. 
First, one can observe that the generated images look very similar to the real images, while being well differentiated in appearance characteristics between both the different classes and the different domains (\eg more red tones in $\mathcal{D}_1$, also with slightly blurry appearance of some classes).
Second, thanks to our distillation during the GAN training, any forgetting is very minor, \ie the generated images $\mathbf{x}'_0$ from $\mathcal{D}_0$ after steps $t$$=$$0$ and $t$$=$$1$ are exhibiting only minute differences.

\subsection{Discussion}
\label{sec:discussion}

\begin{figure*}
    \centering
    \includegraphics[width=\linewidth]{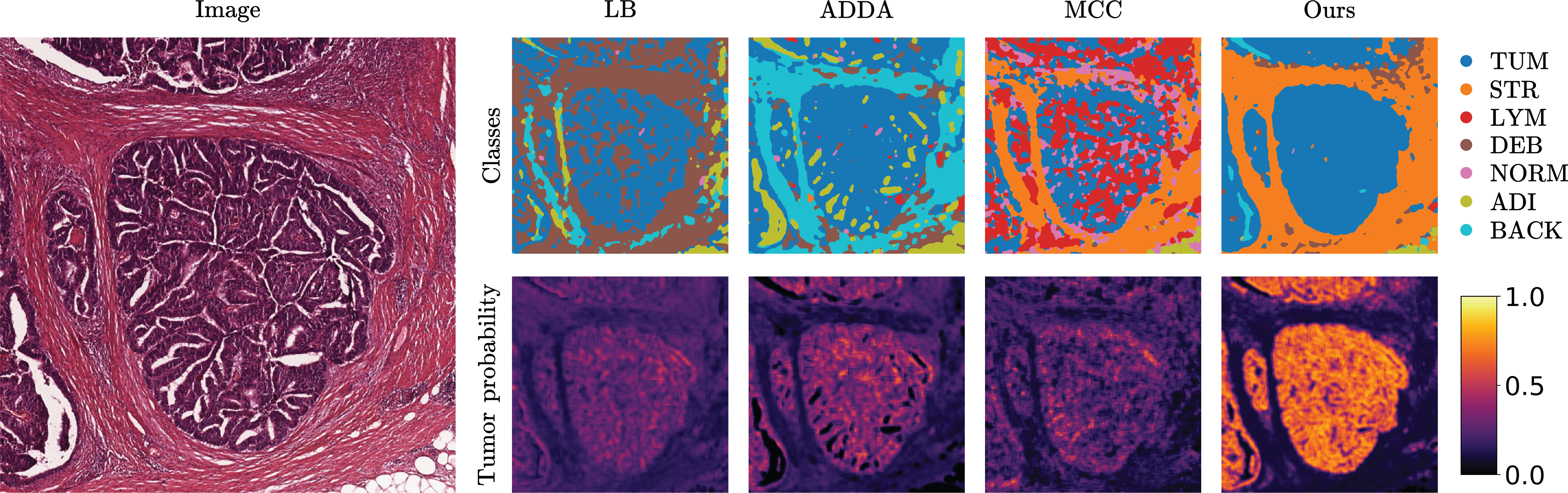}
    \caption{Patch-based segmentation performed by different UDA methods: (left) an original image from $\mathcal{D}_1$; (right) class-wise segmentation maps (top) and predicted probabilities for the tumor (TUM) class (bottom).}
    \label{fig:segmentation}
\end{figure*}

To provide further insights into our approach, we herein compare continual UDA to single-step UDA in \Cref{sec:continual_vs_single}, present ablation experiments of the main method components in \Cref{sec:ablation}, and showcase our method in a patch-based segmentation task for large images in \Cref{sec:segmentation}.

\subsubsection{Continual vs. Single-Step UDA}
\label{sec:continual_vs_single}

In situations where the main focus lies on adaptation to a certain domain, one could simply choose to perform single-step UDA, \eg $\mathcal{D}_0$$\rightarrow$$\mathcal{D}_1$ or $\mathcal{D}_0$$\rightarrow$$\mathcal{D}_2$, instead of continually adapting to multiple target domains.
We show in \Cref{tab:continual_single}, that continual adaptation over multiple target domains can be favorable compared to single-step UDA, even when the intended target is a single domain.
\begin{table}
    \setlength{\tabcolsep}{2pt}
    \caption{Test F1 scores on (a)~$\mathcal{D}_1$ and (b)~$\mathcal{D}_2$, where the models are either only adapted in a single step to one domain (first row) or continually to both target domains (second and third row). All results show the mean $\pm$ std-dev over 5 random initializations.
    Results better than the single-step baseline within the same column are shown in \textbf{bold}.}
    \label{tab:continual_single}
    \begin{subtable}{0.49\linewidth}
        \centering
        \caption{$\mathcal{D}_1$}
        \label{tab:continual_single_d1}
        \begin{tabular}{lc}
        \toprule
        Training & Test F1 ($\%$)\\
        \midrule
        $\mathcal{D}_0$$\rightarrow$$\mathcal{D}_1$ & $86.6\pm0.3$ \\
        $\mathcal{D}_0$$\rightarrow$$\mathcal{D}_1$$\rightarrow$$\mathcal{D}_2$ & $\mathbf{86.8\pm0.7}$ \\
        $\mathcal{D}_0$$\rightarrow$$\mathcal{D}_2$$\rightarrow$$\mathcal{D}_1$ & $82.2\pm0.7$ \\
        \bottomrule
        \end{tabular}
    \end{subtable}
    \hfill
    \begin{subtable}{0.49\linewidth}
        \centering
        \caption{$\mathcal{D}_2$}
        \label{tab:continual_single_d2}
        \begin{tabular}{lc}
        \toprule
        Training & Test F1 ($\%$)\\
        \midrule
        $\mathcal{D}_0$$\rightarrow$$\mathcal{D}_2$ & $64.5\pm1.0$ \\
        $\mathcal{D}_0$$\rightarrow$$\mathcal{D}_1$$\rightarrow$$\mathcal{D}_2$ & $\mathbf{71.9\pm0.8}$ \\
        $\mathcal{D}_0$$\rightarrow$$\mathcal{D}_2$$\rightarrow$$\mathcal{D}_1$ & $\mathbf{70.6\pm0.7}$ \\
        \bottomrule
        \end{tabular}
    \end{subtable}
\end{table}
For both $\mathcal{D}_1$ and $\mathcal{D}_2$, the continual model is seen to outperform the single-step model through at least one long domain sequence. 
Indeed, on $\mathcal{D}_2$, the continual model for either domain sequence performs substantially better than the single-step model.
This striking observation indicates that continual UDA is not only beneficial to train models that perform well across many domains, but it can also serve to increase accuracy in distinct domains, through knowledge sharing across domains.

\subsubsection{Ablation Study}
\label{sec:ablation}

The impact of our proposed components was studied via ablation experiments, with the results reported in \Cref{tab:ablations} for a domain sequence of $\mathcal{D}_0$$\rightarrow$$\mathcal{D}_1$$\rightarrow$$\mathcal{D}_2$.
We first ablate training a continual generator (CG), \ie we only employ generative replay of initial \emph{source} samples, without replaying any synthetic samples from target $\mathcal{D}_1$ when adapting to the next target $\mathcal{D}_2$.
The omission of CG, called ``Ours$-$CG'' in \Cref{tab:ablations}, has the largest impact on the F1 score in $\mathcal{D}_1$, since this is the target domain that is not represented anymore by the samples replayed to the final classifier.
In a second experiment, herein termed ``Ours$-$MFA'', we keep CG but instead replace our proposed MFA with independent discriminators for GAN training and for domain adaptation.
The former operates on the features from the second ResNet block, whereas the latter assesses the features from the last ResNet block.
The results indicate that the lack of MFA affects the performance on all domains similarly, causing a consistent drop  in F1 score of $\approx$4.0\,pp.
This suggests that MFA is important for both adaptation to new domains and mitigation of forgetting, by enabling to employ a common discriminator both for domain adaptation and for GAN training.
Of the two examined components, CG has a bigger impact on the overall performance.

\begin{table}
    \setlength{\tabcolsep}{1.5pt}
    \scriptsize
    \caption{Test F1 scores at the end of $\mathcal{D}_0$$\rightarrow$ $\mathcal{D}_1$$\rightarrow$$\mathcal{D}_2$ for ablations of continual generator (CG) training and multi-scale feature aggregation (MFA).
    The differences compared to our method are shown in parentheses.
    For better readability, mean results are shown without std-dev.}
    \label{tab:ablations}
    \centering
    \begin{tabular}{lllllll}
        \toprule
        \multirow{2}[1]{*}{Training} & \multirow{2}[1]{*}{Method} & \multicolumn{4}{c}{Test F1 (\%)} \\
        \cmidrule{3-6}
        &&$\mathcal{D}_0$ & $\mathcal{D}_1$ & $\mathcal{D}_2$ & overall \\
        \midrule
        \multirow{3}{*}{$\mathcal{D}_0$$\rightarrow$$\mathcal{D}_1$$ \rightarrow$$\mathcal{D}_2$}
        & Ours$-$CG & $81.6$ \diff{-5.9} & $73.4$ \diff{-13.4} & $66.6$ \diff{-5.3} & $73.9$ \diff{-8.2}\\
         & Ours$-$MFA & $83.5$ \diff{-4.0} & $83.0$ \diff{-3.8} & $67.7$ \diff{-4.2} & $78.1$ \diff{-4.0} \\
         & Ours & $87.5$ & $86.8$ & $71.9$ & $82.1$\\
        \bottomrule
    \end{tabular}
\end{table}

\subsubsection{Patch-based Segmentation}
\label{sec:segmentation}

As an extension of classification, our model can also be applied to patch-based segmentation, particularly advantageous for large images.
\Cref{fig:segmentation} shows an example of a 5000$\times$5000 pixel tissue image from $\mathcal{D}_1$ alongside corresponding segmentation maps generated by different methods, \ie LB, ADDA, MCC, and ours.
For each method, we present the segmentation maps computed using the continually trained classifier from the end of the domain sequence $\mathcal{D}_0$$\rightarrow$$\mathcal{D}_2$$\rightarrow$$\mathcal{D}_1$ (as the other sequence yielded very poor and noisy results for the competing methods).
This classifier is applied with a sliding window, classifying its center pixel.
Subsequently, the maps are post-processed by first applying a temperature scaling~\cite{Hinton2015DistillingNetwork} to the logits, computing the softmax, and then employing Gaussian filtering for smoother segmentation maps.
Finally, we also apply a mode filter to reduce salt-and-pepper noise.
While no ground-truth is available for the presented image, our method is seen to produce the most consistent and visually accurate segmentation map among all.
It correctly identifies the large tumor and stromal tissue areas, as well as smaller regions with adipose tissue or plain background.

\section{Conclusions}
We have proposed a novel method for histopathological image classification in a continual UDA setting, in which no images are stored, making our approach applicable in real-world scenarios.
We leverage synergies between GAN training and adversarial domain adaptation by using the same discriminator for both tasks.
We also introduce multi-scale feature aggregation, enabling adaptation to new domains without forgetting previous knowledge.
A potential disadvantage of our method stems from the training challenges inherent to all adversarial UDA approaches, \ie adequate balancing of model capacity between feature extractor and domain discriminator.
However, this can be mitigated, \eg with our proposed early stopping criterion, $R_1$ regularization in the discriminator, and other potential regularization techniques.
We have conducted comprehensive experiments on sequences of colorectal tissue datasets and demonstrate that our approach outperforms existing methods.
We have also conducted ablation experiments confirming the importance of each proposed component.
Furthermore, we have showcased our method for patch-based segmentation of large images.
Since our method is generic, it can be applied also in continual UDA applications with images of various other modalities, \eg from radiology, other microscopy, and photography.

\bibliographystyle{ieeetr}
\bibliography{references}

\end{document}